\documentclass{article}
\usepackage[final]{nips_2017}

\usepackage[utf8]{inputenc} 
\usepackage[T1]{fontenc}    
\usepackage{hyperref}       
\usepackage{url}            
\usepackage{booktabs}       
\usepackage{amsfonts}       
\usepackage{nicefrac}       
\usepackage{microtype}      
\usepackage{graphicx}
\usepackage{xcolor}
\usepackage{csquotes}
\usepackage{enumitem}
\usepackage{wrapfig}
\usepackage{floatrow}
\usepackage{algorithm}
\usepackage{algpseudocode}

\title{Learning model-based planning from scratch}
\author{
  \textbf{Razvan Pascanu\thanks{Denotes equal contribution}, Yujia Li$^\ast$, Oriol Vinyals, Nicolas Heess,}\\
  \textbf{Lars Buesing, Sebastien Racani\`ere, David Reichert,Th\'eophane  Weber,}\\
  \textbf{Daan Wierstra, Peter Battaglia} \\
  DeepMind,
  London, UK\\
  \texttt{\{razp, yujiali, vinyals, heess, lbuesing, sracaniere, \}@google.com} \\
  \texttt{\{ reichert, theophane, wierstra, peterbattaglia\}@google.com}
}

\newcommand{\sa}{s}
\newcommand{\si}{\hat{s}}
\newcommand{\aaa}{a}
\newcommand{\ai}{\hat{a}}
\newcommand{\ra}{r}
\newcommand{\ri}{\hat{r}}

\begin{document}

\maketitle

\begin{abstract}
Conventional wisdom holds that model-based planning is a powerful approach to sequential decision-making. It is often very challenging in practice, however, because  while a model can be used to \textit{evaluate} a plan, it does not prescribe how to \textit{construct} a plan. Here we introduce the ``Imagination-based Planner'', the first model-based, sequential decision-making agent that can learn to construct, evaluate, and execute plans.
Before any action, it can perform a variable number of imagination steps, which involve proposing an imagined action and evaluating it with its model-based imagination. All imagined actions and outcomes are aggregated, iteratively, into a ``plan context'' which conditions future real and imagined actions.
The agent can even decide how to imagine: testing out alternative imagined actions, chaining sequences of actions together, or building a more complex ``imagination tree'' by navigating flexibly among the previously imagined states using a learned policy.
And our agent can learn to plan economically, jointly optimizing for external rewards and computational costs associated with using its imagination.
We show that our architecture can learn to solve a challenging continuous control problem, and also learn elaborate planning strategies in a discrete maze-solving task.
Our work opens a new direction toward learning the components of a model-based planning system and how to use them.
\end{abstract}

\section{Introduction}

Model-based planning involves proposing sequences of actions, evaluating them under a model of the world, and refining these proposals to optimize expected rewards. Several key advantages of model-based planning over model-free methods are that models support generalization to states not previously experienced, help express the relationship between present actions and future rewards, and can resolve states which are aliased in value-based approximations. These advantages are especially pronounced in problems with complex and stochastic environmental dynamics, sparse rewards, and restricted trial-and-error experience.
Yet even with an accurate model, planning is often very challenging because while a model can be used to \textit{evaluate} a plan, it does not prescribe how to \textit{construct} a plan.

Existing techniques for model-based planning are most effective in small-scale problems, often require background knowledge of the domain, and use pre-defined solution strategies. Planning in discrete state and action spaces typically involves exploring the search tree for sequences of actions with high expected value, and apply fixed heuristics to control complexity (e.g., A*, beam search, Monte Carlo Tree Search \cite{coulom2006efficient}). In problems with continuous state and action spaces the search tree is effectively infinite, so planning usually involves sampling sequences of actions to evaluate according to assumptions about smoothness and other regularities in the state and action spaces \cite{busoniu2010reinforcement,hren2008optimistic,munos2011optimistic,weinstein2012bandit}. While most modern methods for planning exploit the statistics of an individual episode, few can learn across episodes and be optimized for a given task. And even fewer attempt to learn the actual planning strategy itself, including the transition model, the policy for choosing how to sample sequences of actions, and procedures for aggregating the proposed actions and evaluations into a useful plan.

Here we introduce the \textit{Imagination-based Planner (IBP)}, a model-based agent which learns from experience all aspects of the planning process: how to construct, evaluate, and execute a plan.
The IBP learns when to act versus when to imagine, and if imagining, how to select states and actions to evaluate which will help minimize its external task loss and internal resource costs. Through training, it effectively develops a planning algorithm tailored to the target problem. The learned algorithm allows it to flexibly explore, and exploit regularities in, the state and action spaces.
The IBP framework can be applied to both continuous and discrete problems. In two experiments we evaluated a continuous IBP implementation on a challenging continuous control task, and a discrete IBP in a maze-solving problem.

Our novel contributions are:
\begin{itemize}[noitemsep,nolistsep]
    \item A fully learnable model-based planning agent for continuous control.
    \item An agent that learns to \textit{construct} a plan via model-based imagination.
    \item An agent which uses its model of the environment in two ways: for imagination-based planning and gradient-based policy optimization.
    \item A novel approach for learning to build, navigate, and exploit ``imagination trees''.
\end{itemize}

\subsection{Related work}
Planning with ground truth models has been studied heavily and led to remarkable advances. AlphaGo \cite{silver2016mastering}, the world champion computer Go system, trains a policy to decide how to expand the search tree using a known transition model.
Planning in continuous domains with fixed models must usually exploit background knowledge to sample actions efficiently \cite{busoniu2010reinforcement,hren2008optimistic,munos2011optimistic}. Several recent efforts \cite{watter2015embed,lenz2015deepmpc,finn2017deep} addressed planning in complex systems, however, the planning itself uses classical methods, e.g., Stochastic Optimal Control, trajectory optimization, and model-predictive control.

There have also been various efforts to learn to plan. The classic ``Dyna'' algorithm learns a model, which is then used to train a policy \cite{sutton1991dyna}. Vezhnevets et al. \cite{vezhnevets2016strategic} proposed a method that learns to initialize and update a plan, but which does not use a model and instead directly maps new observations to plan updates. The value iteration network \cite{Tamar2016} and predictron \cite{silver2016predictron} both train deep networks to implicitly plan via iterative rollouts. However the former does not use a model, and the latter uses an abstract model which does not capture the world dynamics, and was only applied to learning Markov reward processes rather than solving Markov decision processes (MDPs).

Our approach is also related to classic work on meta-reasoning \cite{Russel1991,Horvitz1988,Hay2012}, in which an internal MDP schedules computations, which carry costs, in order to solve a task. More recently, neural networks have been trained to perform ``conditional'' and ``adaptive computation'' \cite{Bengio2013,Bengio2015,graves2016adaptive}, which results in a dynamic computational graph.

Recently \cite{Fragkiadaki2015} trained a ``visual imagination'' model to control simulated billiards systems, though their system did not learn to plan.
Our IBP was most inspired by Hamrick et al.'s ``imagination-based metacontroller'' (IBMC) \cite{hamrick2017metacontrol}, which learned an adaptive optimization policy for one-shot decision-making in contextual bandit problems. Our IBP, however, learns an adaptive planning policy in the more general and challenging class of sequential decision-making problems.Similar to our work is the study of \cite{I2A} that looks in detail at dealing with imperfect complex models of the world, working on pixels, in the discrete sequential decision making processes.

\begin{figure}[t]
\floatbox[{\capbeside\thisfloatsetup{capbesideposition={right,top},capbesidewidth=0.55\textwidth}}]{figure}[\FBwidth]
{\caption{\textbf{IBP schematic.} The manager (blue circle) chooses whether to imagine (top row) or act (bottom row). Imagining consists of the controller (red square) proposing an action and the imagination (green cloud icon) evaluating it. Acting consists of the controller proposing an action, which is executed in the world (Earth icon). The memory (yellow triangle) aggregates information from this iteration and updates the history. }\label{fig:summary-schematic}}
{\includegraphics[width=0.45\textwidth]{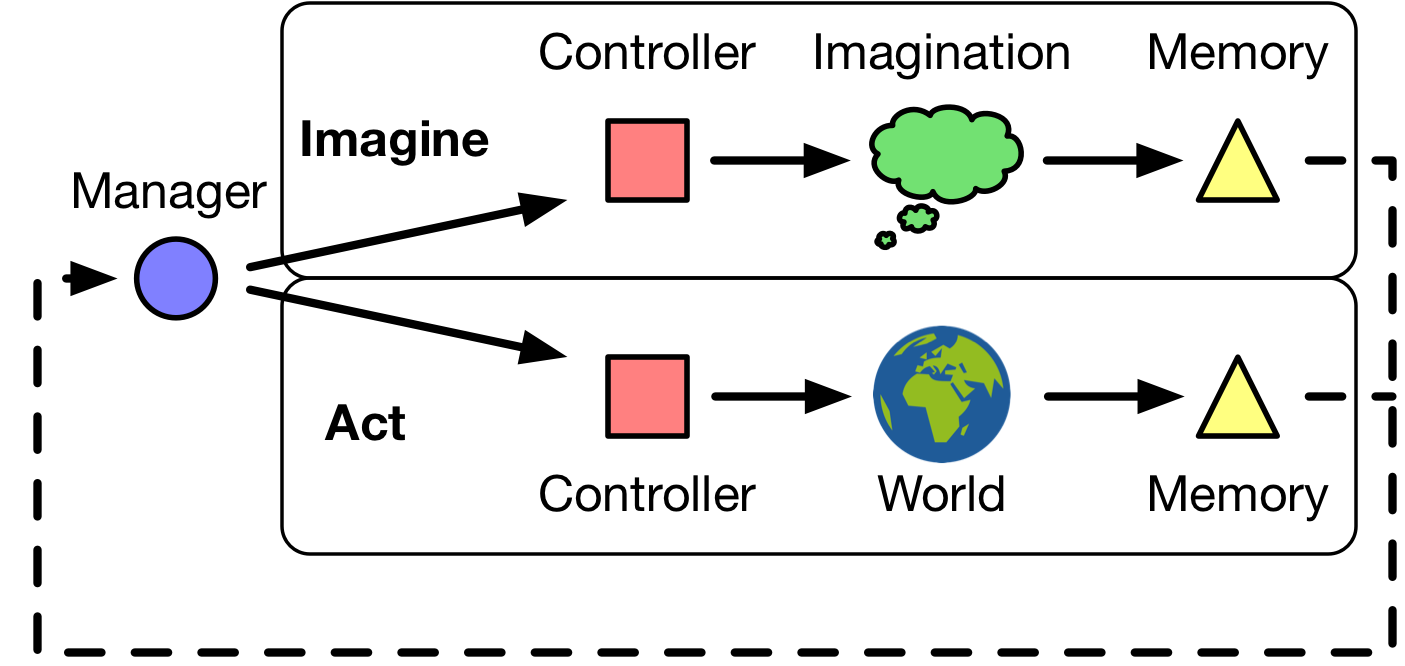}}
\end{figure}

\begin{figure}[t]
  \centering
  \includegraphics[width=0.95\textwidth]{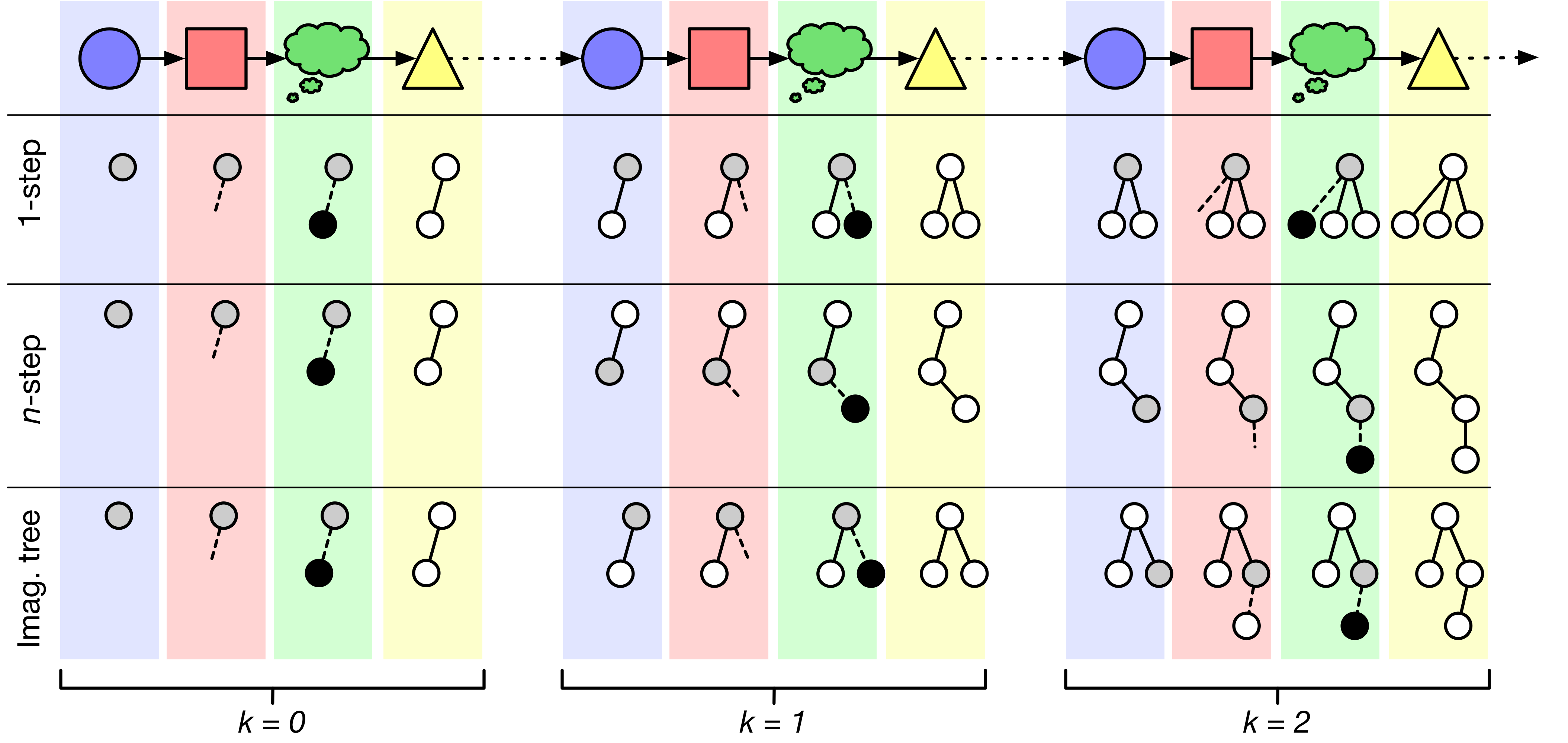}
  \caption{\textbf{Imagination strategies.} Sequences of imagined actions (edges) and states (nodes) form a tree, over $k=0, 1, 2$ imagination steps. Grey nodes indicate states chosen by the manager (blue circle) from which to imagine, dotted edges indicate new actions proposed by the controller (red square), and black nodes indicate future states predicted by the imagination (green cloud icon). The three imagination strategies correspond to the three rows of trees being constructed (see text). }
  \vspace{-8pt}
  \label{fig:tree-types}
\end{figure}

\vspace{-8pt}
\section{Model}
\vspace{-8pt}

\label{sec:compute_descr}

The definition of planning we adopt here involves an agent proposing sequences of actions, evaluating them with a model, and following a policy that depends on these proposals and their predicted results. Our IBP implements a recurrent policy capable of planning via four key components (Figure~\ref{fig:summary-schematic}). On each step, the \textit{manager} chooses whether to imagine or act. If acting, the \textit{controller} produces an action which is executed in the world. If imagining, the controller produces an action which is evaluated by the model-based \textit{imagination}. In both cases, data resulting from each step are aggregated by the \textit{memory} and used to influence future actions. The collective activity of the IBP's components supports various strategies for constructing, evaluating, and executing a plan.

On each iteration, $i$, the IBP either executes an action in the world or imagines the consequences of a proposed action. The actions executed in the world are indexed by $j$, and the sequence of imagination steps the IBP performs before an action are indexed by $k$. Through the planning and acting processes, two types of data are generated: external and internal. External data includes observable world states, $\sa_j$, executed actions, $\aaa_j$, and obtained rewards, $\ra_j$. Internal data includes imagined states, $\si_{j,k}$, actions, $\ai_{j,k}$, and rewards, $\ri_{j,k}$, as well as the manager's decision about whether to act or imagine (and how to imagine), termed the ``route'', $u_{j,k}$, the number of actions and imaginations performed, and all other auxiliary information from each step. We summarize the external and internal data for a single iteration as $d_i$, and the history of all external and internal data up to, and including, the present iteration as, $h_i = (d_0, \dots, d_i)$. The set of all imagined states since the previous executed action are $\{\si_{j,0}, \dots, \si_{j,k}\}$, where $\si_{j,0}$, is initialized as the current world state, $\sa_j$.

The \textit{manager}, $\pi^M: \mathcal{H} \rightarrow \mathcal{P}$, is a discrete policy which maps a history, $h \in \mathcal{H}$, to a route, $u \in \mathcal{U}$. The $u$ determines whether the agent will execute an action in the environment, or imagine the consequences of a proposed action. If imagining, the route can also select which previously imagined, or real, state to imagine from. We define $\mathcal{U} = \{\mathrm{act}, \si_{j,0}, \dots, \si_{j,k}\}$, where $\mathrm{act}$ is the signal to act in the world, and the $\si_{j,l}$ signals to propose and evaluate an action from imagined state, $\si_{j,l}$.

The \textit{controller}, $\pi^C: \mathcal{S} \times \mathcal{H} \rightarrow \mathcal{A}$, is a contextualized action policy which maps a state $s \in \mathcal{S}$ and a history to an action, $a \in \mathcal{A}$. The state which is provided as input to the controller is determined by the manager's choice of $u$. If executing, the actual state, $\sa_j$, is always used. If imagining, the state $\si_{j,l}$ is used, as mentioned above. There are different possible imagination strategies, detailed below, which determine which state is used for imagination.

The \textit{imagination}, $I: \mathcal{S} \times \mathcal{A} \rightarrow \mathcal{S} \times \mathcal{R}$ is a model of the world, which maps states, $s \in \mathcal{S}$, and actions, $a \in \mathcal{A}$, to consequent states, $s' \in \mathcal{S}$, and scalar rewards, $r \in \mathcal{R}$.

The \textit{memory}, $\mu: \mathcal{D} \times \mathcal{H} \rightarrow \mathcal{H}$, recurrently aggregates the external and internal data generated from one iteration, $d \in \mathcal{D}$, to update the history, i.e., $h_i = \mu(d_i, h_{i-1})$.

Constructing a plan involves the IBP choosing to propose and imagine actions, and building up a record of possible sequences of actions' expected quality. If a sequence of actions predicted to yield high reward is identified, the manager can then choose to act and the controller can produce the appropriate actions.

We explored three distinct imagination-based planning strategies: ``$1$-step'', ``$n$-step'', and ``tree'' (see Figure~\ref{fig:tree-types}). They differ only by how the manager selects, from the actual state and all imagined states since the last action, the state from which to propose and evaluate an action.
For $1$-step imagination, the IBP must imagine from the actual state, $\si_{j,0}$. This induces a depth-$1$ tree of imagined states and actions (see Figure~\ref{fig:tree-types}, first row of graphs).
For $n$-step imagination, the IBP must imagine from the most recent previously imagined state, $\si_{j,k}$. This induces a depth-$n$ chain of imagined states and actions (see Figure~\ref{fig:tree-types}, second row of graphs).
For trees, the manager chooses whether to imagine from the actual state or any previously imagined state since the last actual action, $\{\si_{j,0}, \dots, \si_{j,k}\}$. This induces an ``imagination tree'', because imagined actions can be proposed from any previously imagined state (see Figure~\ref{fig:tree-types}, third row of graphs).

\section{Experiment 1: Continuous control}

\subsection{Spaceship task}
We evaluated our model in a challenging continuous control task adapted from the ``Spaceship Task'' in \cite{hamrick2017metacontrol}, which we call ``Spaceship Task 2.0'' (see Figure~\ref{fig:trajectories} and videos URL: \href{https://drive.google.com/open?id=0B3u8dCFTG5iVaUxzbzRmNldGcU0}{https://drive.google.com/open?id=0B3u8dCFTG5iVaUxzbzRmNldGcU0}). The agent must pilot an initially stationary spaceship in 2-D space from a random initial position at a radius between $0.6$ and $1$ distance units, and a random mass between $0.004$ and $0.36$ mass units, to its mothership, which is always set to position $(0,0)$. The agent can fire its thrusters with a force 
of its choice, which accelerates its velocity by $F = m a$. There are five stationary planets in the scene, at random positions at a radius between $0.4$ and $1$ distance units
and with masses that vary between $0.08$ and $0.4$ mass units. The planets' gravitational fields accelerate the spaceship, which induces complex, non-linear dynamics. A single action entailed the ship firing its thrusters on the first time step, then traveling under ballistic motion for 11 time steps under the gravitational dynamics.

There are several other factors that influence possible solutions to the problem. The spaceship pilot must pay a linearly increasing price for fuel ($\mathrm{price} = 0.0002 \mbox{ or } 0.0004$ cost units), when the force magnitude is greater then a threshold value of $8$ distance units, i.e., $\mathrm{fuel\_cost} = \max(0, (\mathrm{force\_magnitude} - 8) \cdot \mathrm{price}$). This incentivizes the pilot to choose small thruster forces. We also included multiplicative noise in the control, which further incentivizes small controls and also bounds the resolution at the future states of the system can be accurately predicted.

\subsection{Neural network implementation and training}

We implemented our continuous IBP for the spaceship task using standard deep learning building blocks.
The memory, $\mu$, was an LSTM \cite{hochreiter1997long}.
Crucially, the way the continuous IBP encodes a plan is by embedding $h_i$ into a ``plan context'', $c_i$, using $\mu$.
Here the inputs to $\mu$ were the concatenation of a subset of $h_i$. For imagining, they were: $(p_{j,k}, \sa_j, \si_{j,p_{j,k}}, \ai_{j,k}, \si_{j,k+1}, \ri_{j,k}, j, k, c_{i-1})$. For acting, they were: $(p_{j,k}, \sa_j, \si_{j,0}, \aaa_j, \sa_{j+1}, \ra_j, j, k, c_{i-1})$.
The manager, $\pi^M$, and controller, $\pi^C$, were multi-layer perceptrons (MLP). The $\pi^M$ took $\sa_j$ and $c_{i-1}$ as input, and output $p_{j,k}$. The $\pi^C$ took $\si_{j,p_{j,k}}$ and $c_{i-1}$ as input, and output $\ai_{j,k}$ or $\aaa_j$, for imagining or acting, respectively.
And the imagination-based model of the environment, $I$ was an interaction network (IN) \cite{battaglia2016interaction}, a powerful neural architecture to model graph-like data. It took as input $(\si_{j,k}, \ai_{j,k})$ and returned $\si_{j,k+1}$ for imagining, and took $(\sa_j, \aaa_j)$ and returned $\sa_{j+1}$ for acting.

The continuous IBP was trained to jointly optimize two loss terms, the external \textit{task} and internal \textit{resource} losses. The task loss reflects the cost of executing an action in the world, including the fuel cost and final $L2$ distance to the mothership.
The resource loss reflects the cost of using the imagination on a particular time step and only affects the manager.
It could be fixed across an episode, or vary with the number of actions taken so far, expressing the constraint that imagining early is less expensive than imagining on-the-fly. The total loss that was optimized was the sum of the task and resource losses.

The training consisted of optimizing, by gradient descent (Adam \cite{kingma2014adam}), the parameters of the agent with respect to the task loss and internal resource costs. The computational graph of the IBP architecture was divided into three distinct subgraphs: 1) the model, 2) the manager, and 3) the controller and memory. Each subgraph's learnable parameters were trained simultaneously, on-policy, but independently from one another.

The $I$ model was trained to make next-step predictions of the state in a supervised fashion, with error gradients computed by backpropagation. The data was collected from the observations the agent makes when acting in the real world.
Because $\pi^M$'s outputs were discrete, we used the REINFORCE algorithm \cite{williams1992simple} to estimate its gradients. Its policy was stochastic, so we also applied entropy regularization during training to encourage exploration. The rewards for the manager consist of the negative sum between the internal and external loss (the cost of each step is payed at each step, while the $L_2$ loss is payed at the end of the sequence).

The $\pi^C$ and $\mu$ were trained jointly. Because the $\pi^M$'s output was non-differentiable, we treated the routes it chose for a given episode as constants during learning. This induced a computational graph which varied as a function of the route value, where the route was a switch that determined the gradients' backward flow. In order to approximate the error gradient with respect to an action executed in the world, we used stochastic value gradients (SVG) \cite{heess2015learning}. We unrolled the full recurrent loop of imagined and real controls, and computed the error gradients using backpropagation through time. $\pi^C$ and $\mu$ are trained only to minimize the external loss, i.e. the fuel cost and the final L2 distance to the mothership, but not including the imagination cost. The training regime is similar to the one used in \cite{hamrick2017metacontrol}.

In terms of strategies that we relied on, beside \textit{1-step} and \textit{n-step} we used only a restricted version of the imagination-tree strategy where the manager only
selected from $\{\mathrm{act}, \si_{j,0}, \si_{j,k}\}$ as depicted in Figure~\ref{fig:tree-types}. While this reduced the range of trees the model can construct, it was presumably easier to train because it had a fixed number of discrete route choices, independent of $k$. Future work should explore using RNNs or local navigation within the tree to handle variable sized sets of route choices for the manager.

\subsection{Results}

\begin{figure}[t]
  \centering
  \begin{tabular}{ccc}
  \includegraphics[width=.87\textwidth, trim=0 1in 0 1in,clip]{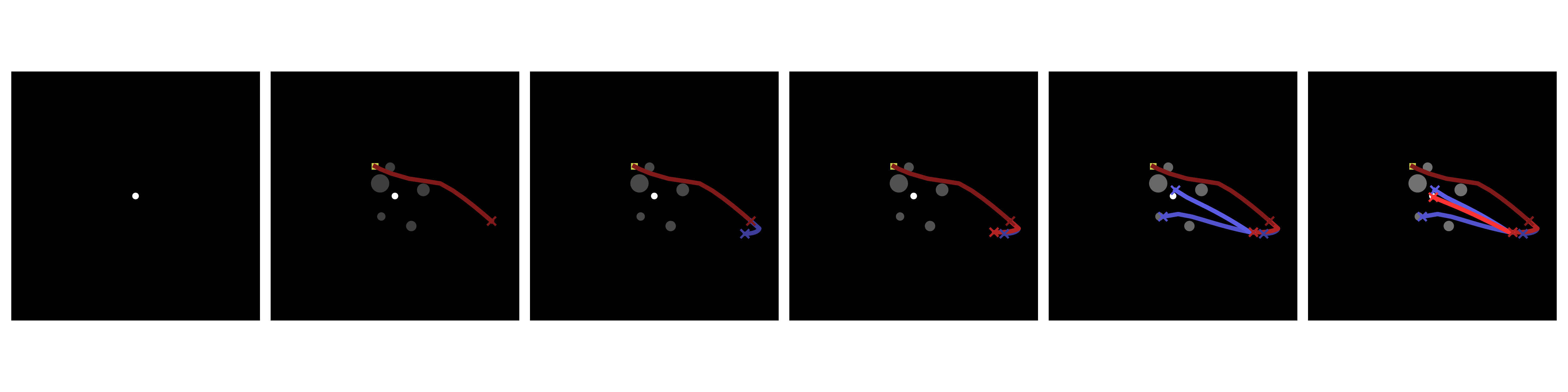} &
   \includegraphics[height=.12\textwidth]{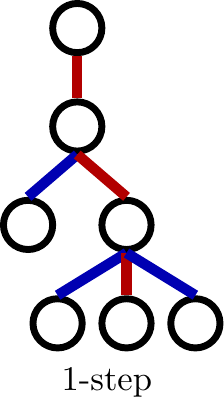} \\
  \includegraphics[width=.87\textwidth, trim=0 1in 0 1in,clip]{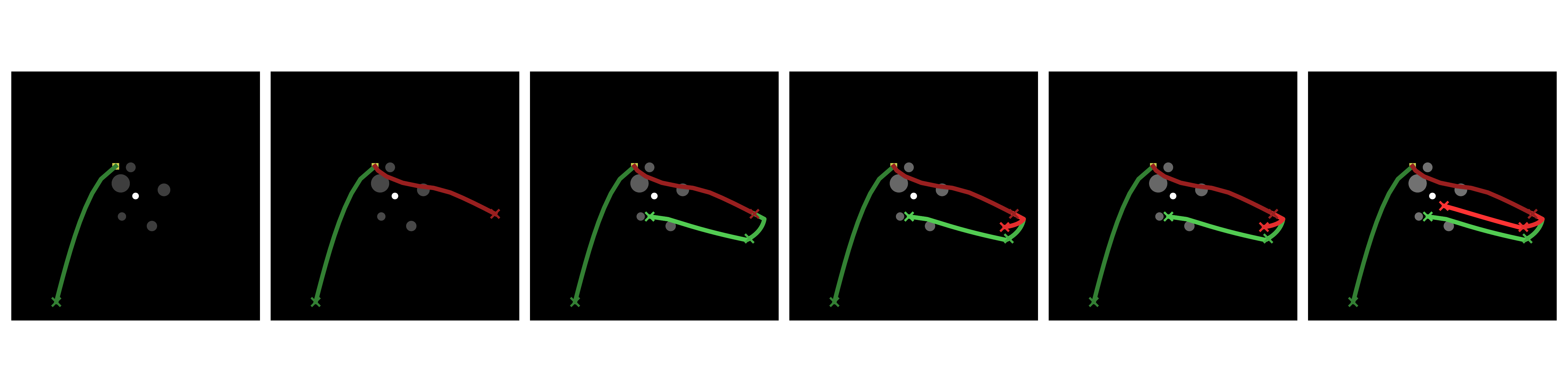} &
  \includegraphics[height=.12\textwidth]{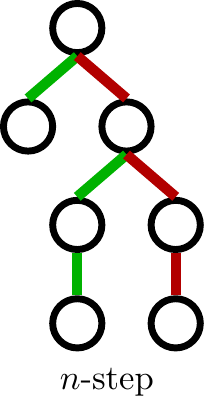} \\
  \includegraphics[width=.87\textwidth, trim=0 1in 0 1in,clip]{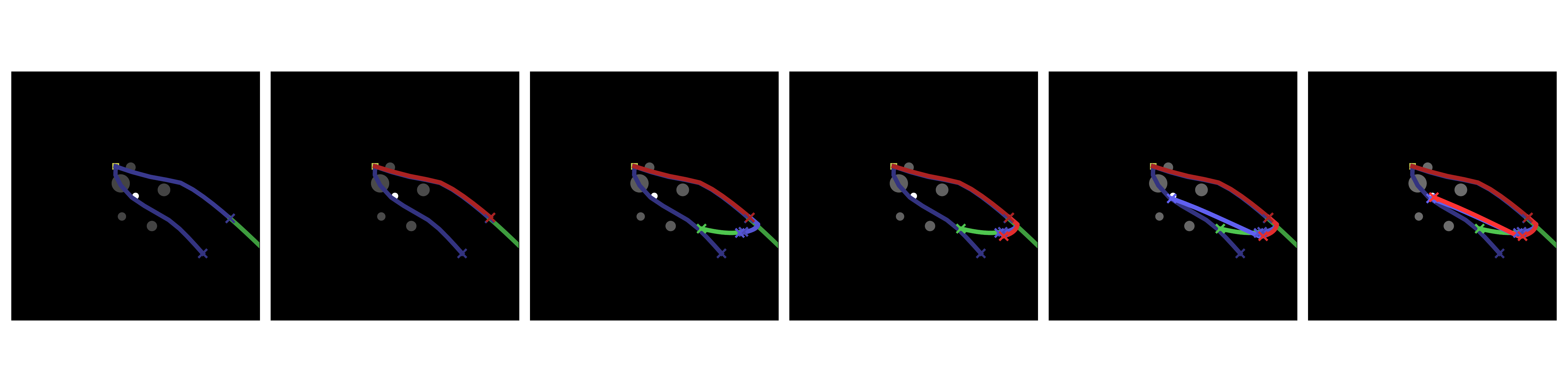} &
  \includegraphics[height=.12\textwidth]{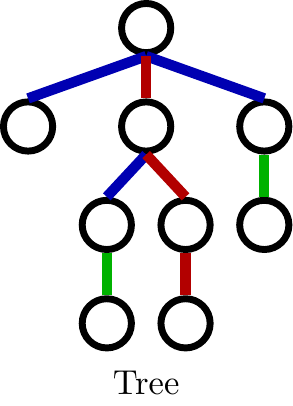} \\
  \includegraphics[width=.87\textwidth, trim=0 1in 0 1in,clip]{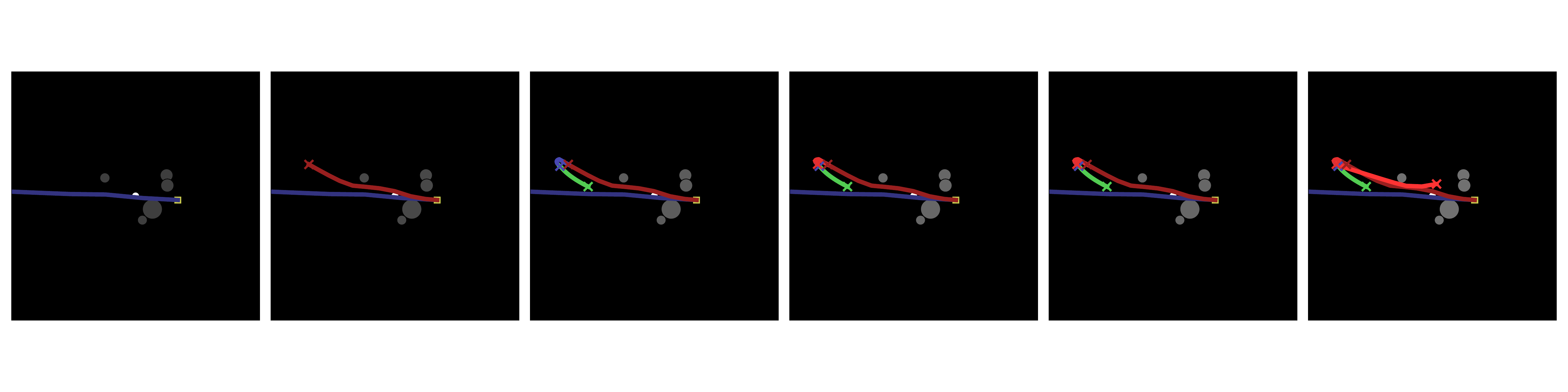}&
  \includegraphics[height=.12\textwidth]{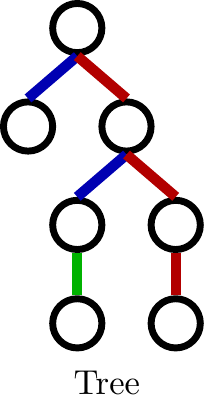}
  \end{tabular}

  \caption{\textbf{Plans constructed by the agent on four different episodes}.
  The images from left to right imagined and executed action rollouts over time. Each step covers two panels, the first one only the imagination (if any) while the second one contains an executed action.
  Note that real actions are depicted with red, and imagination steps with either blue or green. Blue actions start from the current world state. Green start from previously imagined states (for $n$-step strategy all arrows are green). The right-most column shows the tree representing all real actions and imaginations executed by the agent.}
  Videos URL: \href{https://drive.google.com/open?id=0B3u8dCFTG5iVaUxzbzRmNldGcU0}{https://drive.google.com/open?id=0B3u8dCFTG5iVaUxzbzRmNldGcU0}
  \label{fig:trajectories}
\end{figure}

\begin{figure}[t]
  \centering
  \includegraphics[width=.8\textwidth]{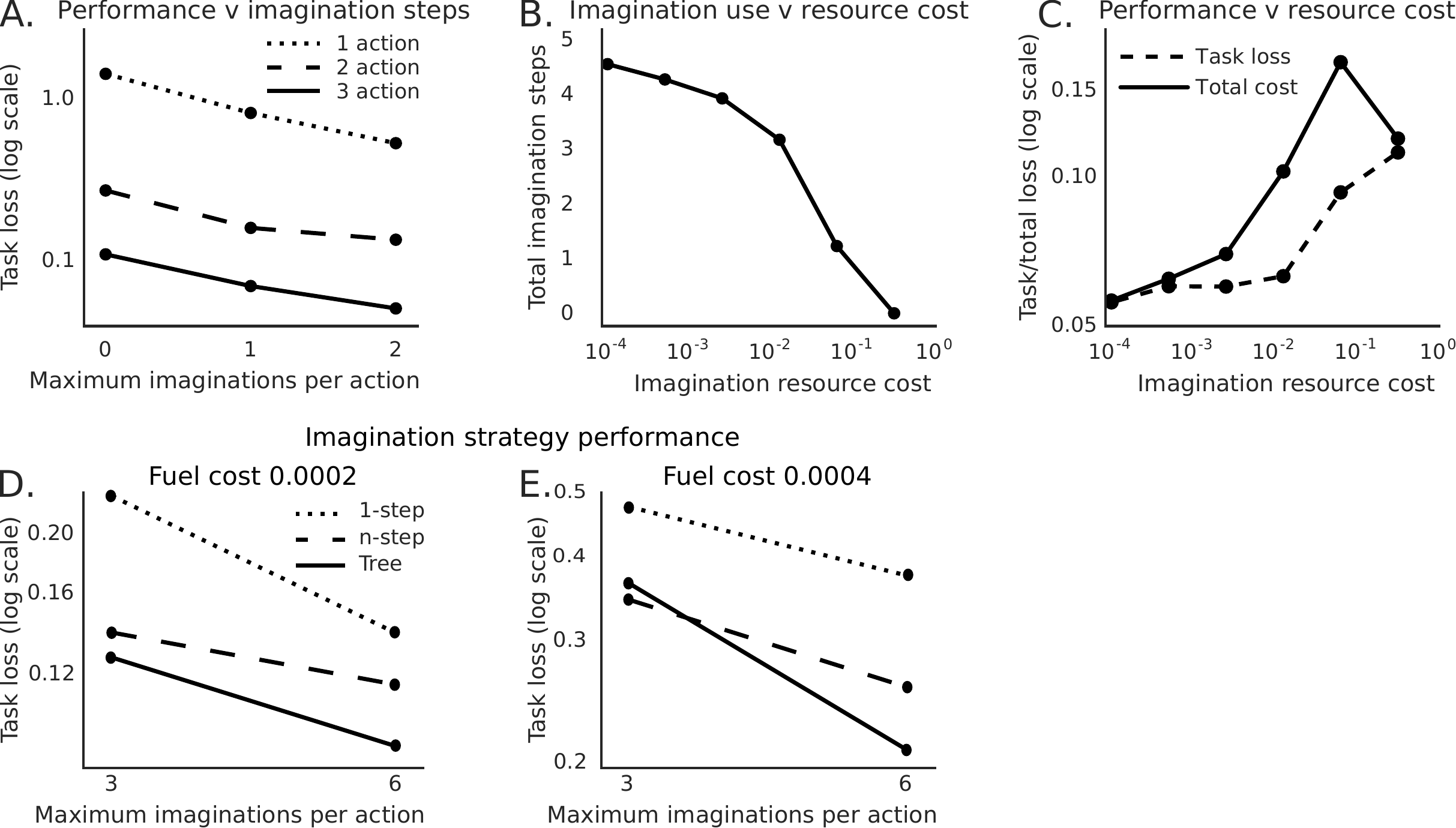}
  \caption{\textbf{Spaceship task performance} \textbf{a. Task loss as a function of numbers of imaginations and actions.} The x-axis represents the maximum number of imagination steps. The y-axis represents the average task loss (log-scale). The dotted, dashed, and solid lines represent agents which are allowed 1, 2, and 3, actions, respectively. There was zero resource cost applied to the use of imagination. \textbf{b. Use of imagination as a function of resource cost.} The agent here (and in c.) was allowed three actions, and a maximum of two imaginations per action. The x-axis represents the internal resource cost on imagination, $\tau$. The y-axis represents the average number of imagination steps the agent chose to use per episode (maximum of 6). \textbf{c. Task loss as a function of resource cost.} The x-axis represents the internal resource cost on imagination, $\tau$. The y-axis represents the average task loss (log-scale). The dashed and solid lines represent the external task loss and the total (task plus resource) loss, respectively. \textbf{d-e. Task loss as a function of numbers of imaginations and imagination strategy.} The x-axis represents the maximum number of imagination steps. The y-axis represents the average task loss (log-scale). The dotted, dashed, and solid lines represent agents which use the $1$-step, $n$-step, and imagination-tree strategies, respectively. There was increasing resource cost applied to the use of imagination, starting from paying nothing, and increasing by 0.5 after each execution. (d) and (e) differ by the fuel costs, (e) being more challenging. }
  \label{fig:performance-spaceship}
\end{figure}

Our first results show that the $1$-step lookahead IBP can learn to use its model-based imagination to improve its performance in a complex, continuous, sequential control task. Figure~\ref{fig:trajectories}'s first row depicts several example trajectories from a $1$-step IBP that was granted three external actions and up to two imagined actions per external action (a maximum of nine imagined and external actions). 
The IBP learned to often use its first two external actions to navigate to open regions of space, presumably to avoid the complex nonlinear dynamics imposed by the planets' gravitational fields, and then use its final action to seek the target location. It used its imagined actions to try out potential actions, which it refined iteratively. Figure~\ref{fig:performance-spaceship}a shows the performance of different IBP agents, which are granted one, two, and three external actions, and different maximum numbers of imagined actions.
The task loss always decreased as a function of the maximum number of imagined actions. The version which was allowed only one external action (blue line) roughly corresponded to Hamrick et al.'s \cite{hamrick2017metacontrol} IBMC. The IBP agents that could not use their imagination (left-most points, 0 imaginations per action)
represents the SVG baselines on this domain: they could use their model to compute policy gradients during training, but could not use their model to evaluate proposed actions. These results provide clear evidence of the value of imagination for sequential decision-making, even with only 1-step lookahead.

We also examined the $1$-step lookahead IBP's use of imagination as greater resource costs, $\tau$, were imposed (Figure~\ref{fig:performance-spaceship}b), and found that it smoothly decreased the total number of imagination steps to avoid this additional penalty, eventually using no imagination at all when under high values of $\tau$. The reason the low-cost agents (toward the left side of Figure~\ref{fig:performance-spaceship}b) did not use the full six imaginations allowed is because of the small entropy reward, which incentivized it to learn a policy with increased variance in its route choices. Figure~\ref{fig:performance-spaceship}c shows that the result of increased imagination cost, and decreased use of imagination was that the task loss increased.

We also trained an $n$-step IBP in order to evaluate whether the IBP could construct longer-term plans, as well as a restricted imagination-tree IBP to determine whether it could learn more complex plan construction strategies. Figure~\ref{fig:trajectories}'s second row shows $n$-step IBP trajectories, and the third and fourth rows show two imagination-tree IBP trajectories. In this praticular task, after each execution the cost of an imagination step increases, making it more advantageous to plan early.  Our results (Figure~\ref{fig:performance-spaceship}d-e) show that $n$-step IBP performance is better than $1$-step, and that an imagination-tree IBP outperforms both, especially when more maximum imagination steps are allowed, for two values of the fuel cost we applied. Note that imagination-tree IBP becomes more useful as the agent has more imagination steps to use.

\section{Experiment 2: Discrete mazes}

In the second, preliminary experiment, a discrete 2D maze-solving task, we implemented a discrete IBP to probe its use of different imagination strategies. For simplicity, the imagination used a perfect world model. The controller is given by a tabular representation, and the history represents incremental changes to it, caused by imagination steps. The memory's aggregation functionality was implemented by accumulating the updates induced by imagination steps into the history. The controller's policy was determined by the sum of
the learned Q-value and the history tensor. The manager was a convolutional neural network (CNN) that took as input the current Q table, history tensor, and map layout.

In this particular instance we explored IBP as a search strategy.
We created mazes for which states would be aliased in the tabular Q representation. Each maze could have multiple candidate goal locations, and in each episode a goal location was selected at random from the candidates. An agent was instantiated to use a single set of Q-values for each maze and all its different goal locations.
This would also allow a model-based planning agent to use imagined rollouts to disambiguate states and generalize to goal locations unseen during training.

More results on these maze tasks can be found in the supplementary material.

\begin{figure}[h]
\centering
\includegraphics[width=0.7\textwidth]{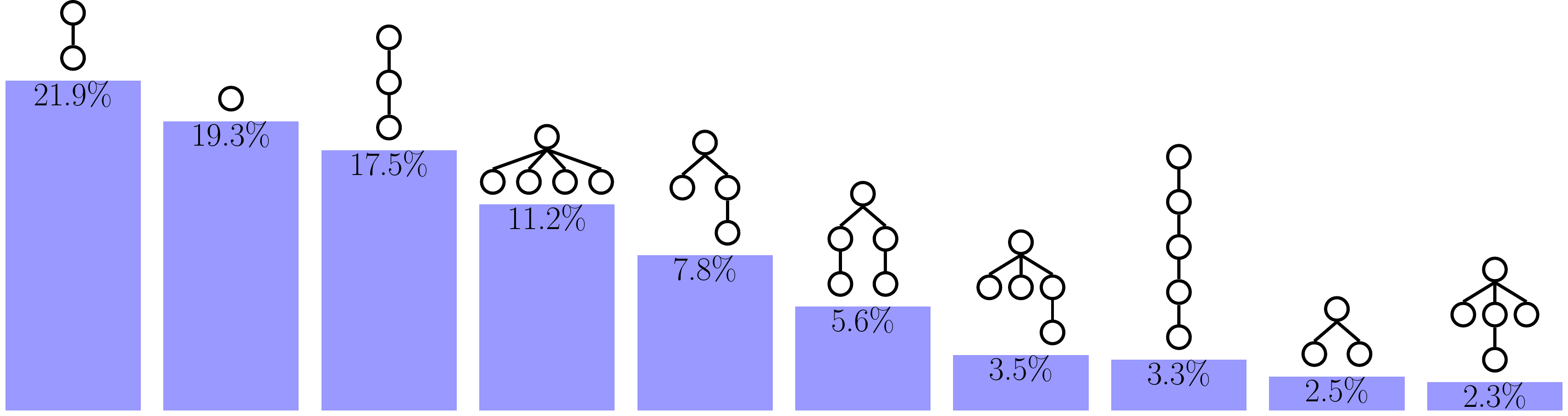}
\caption{Learned imagination trees for the maze task.  See supplementary material for more details.}
\label{fig:maze-tree-histogram}
\end{figure}

\begin{figure}
  \hspace{-7pt}
  \begin{tabular}{cccccc}
  &
      \includegraphics[width=0.14\textwidth]{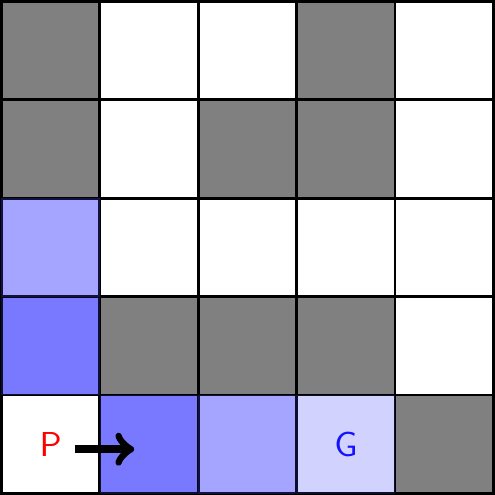} &
      \includegraphics[width=0.14\textwidth]{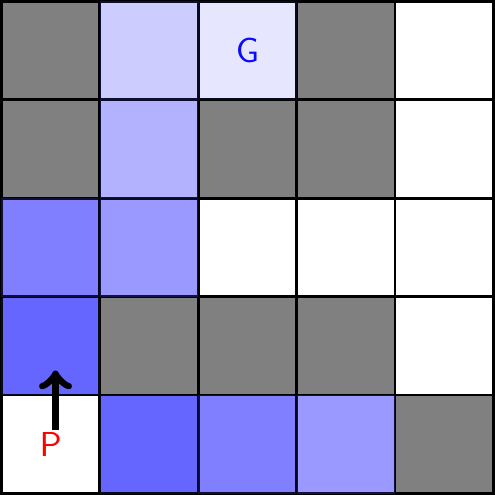} &
      \includegraphics[width=0.14\textwidth]{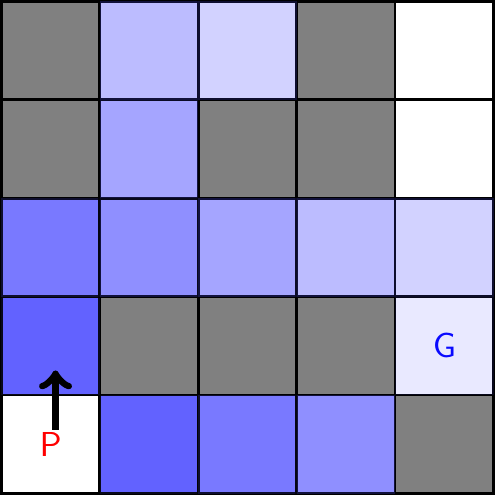} &
      \includegraphics[width=0.14\textwidth]{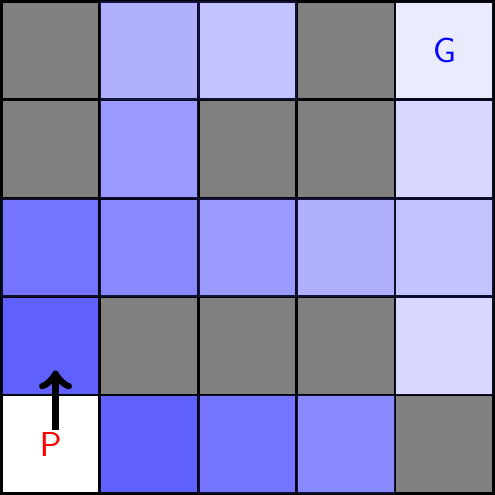} & \\
        \includegraphics[width=0.14\textwidth]{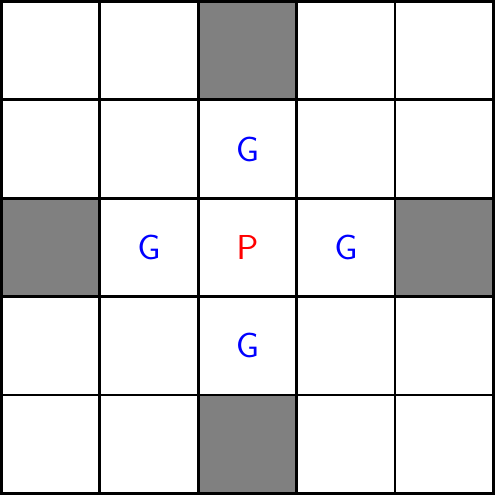} &
    \includegraphics[width=0.14\textwidth]{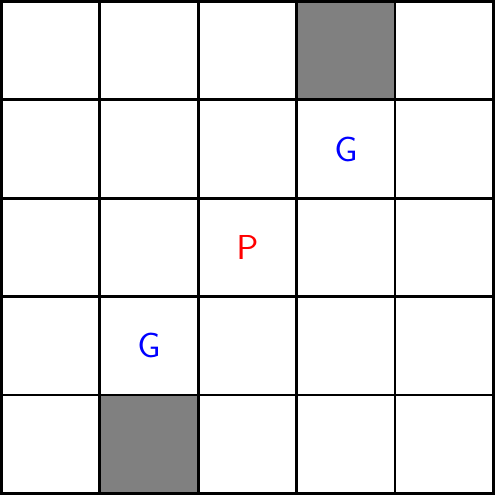} &
    \includegraphics[width=0.14\textwidth]{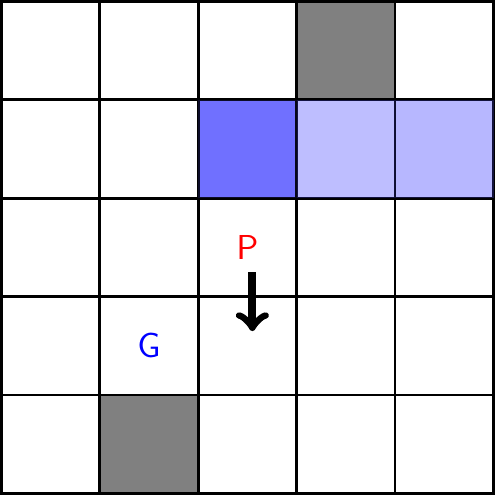} &
    \includegraphics[width=0.14\textwidth]{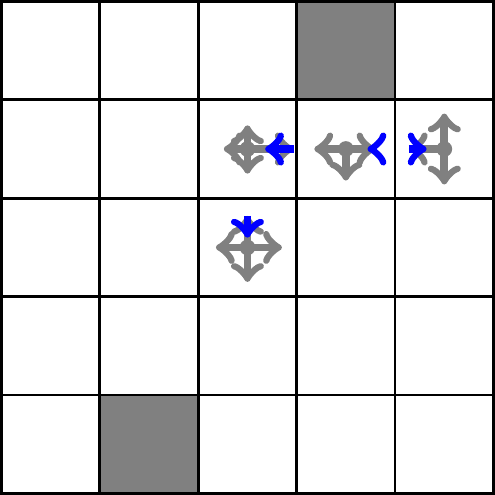} &
    \includegraphics[width=0.14\textwidth]{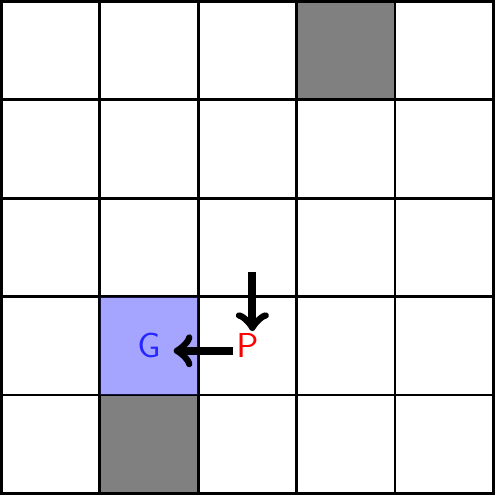} &
    \includegraphics[width=0.14\textwidth]{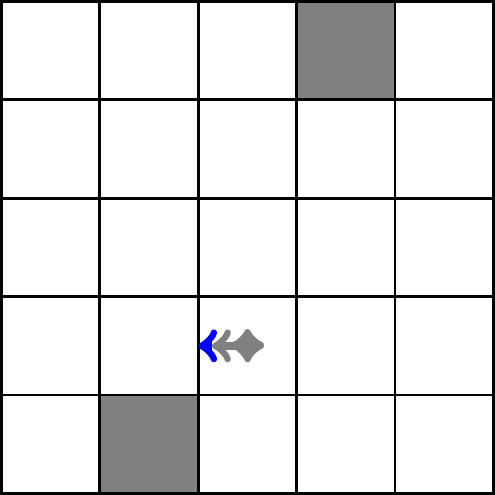}
  \end{tabular}
  \caption{\textbf{Top row}: Single maze, multiple goals, {\color{red} P} represents the position of the player, and {\color{blue}G} is the goal.
  The first three configurations were from training, and the last tested the agent's ability to generalize.  The imagination steps are the shaded blue areas, for different imagination budgets 5, 9, 13 and 15 from left to right.  The imagination depth corresponds to the color saturation of the grid cells.  The actual action the agent took is shown as a solid black arrow. \textbf{Bottom row:} Multiple mazes and one example run.  Left two: two example mazes, one with more candidate goal locations than the other.  Middle and right four: one example run of a learned agent on one maze.  In this run, the agent performed four imagination steps, then took its first action, then performed one more imagination step.  Both the imagination steps and the rollout updates to $c$ are shown.  In the rollout updates, the original $Q+c$ values before the rollout are colored gray, and the updates are blue.}
  \label{fig:maze}
\end{figure}

\textbf{Single Maze Multiple Goals:} We start by exploring state aliasing and generalization to out-of-distribution states. Figure~\ref{fig:maze} top row shows the prototypical setup we consider for this exploration, along with imagination rollout examples.  During training, the agent only saw the first three goal positions, selected at random, but never the fourth.
The reward was -1 for every step, except at the end, when the agent received as reward the number of steps left in its total budget of 20 steps.

The amount of available actions at each position in the maze was small, so we limited the manager to a fixed policy that always expanded the next leaf node in the search tree which had the highest value.  Figure~\ref{fig:maze} also showed the effect of having different number of imaginations to disambiguate the evaluated maze. With sufficient imagination steps, not only could the agent find all possible goals, but most importantly, it could resolve new goal positions not experienced during training.

\textbf{Multiple Mazes Multiple Goals: }
Here we illustrate how a learned manager adapts to different environments and proposes different imagination strategies.  We used different mazes, some with more possible goal positions than others, and with different optimal strategies. We introduced a regularity that the manager could exploit: we made the wall layouts be correlated with the number of possible goal positions.
The IBP learned a separate tabular control policy for each maze, but used a shared manager across all mazes.

Figure~\ref{fig:maze}'s bottom row shows several mazes and one example run of a learned agent. Note, the maze with two goals cannot be disambiguated using just $1$-step imagination--longer imagination sequences are needed. And the $n$-step IBP will have trouble resolving the maze with four goals in a small number of steps because it cannot jump back and check for a different hypothesized goal position.  The imagination-tree IBP could adapt its search strategy to the maze, and thus achieved an overall average reward that was roughly 25\% closer to the optimum than either the $1$-step or $n$-step IBPs.

This experiment also exposes a limitation of the simple imagination-tree strategy, where the manager can only choose to imagine only from $\{\si_{j,0}, \si_{j,k}\}$. It can only proceed with imagining or reset back to the current world state. So if a different path should be explored, but part of its path overlaps with a previously explored path, the agent must waste imagination steps in reconstructing this overlapping segment.

One preliminary way to address this is to allow the manager to work with fixed-length macro actions done by the control policy, which effectively increases the imagination trees' branching factor while making the tree more shallow. In Supplementary Materials we show the benefits of using macros on a scaled up (7$\times$7) version of the tasks considered in Figure~\ref{fig:maze}.  Figure~\ref{fig:maze-tree-histogram} shows the most common imagination trees constructed by the agent, highlighting the diversity of the IBP's learned planning strategies.

\section{Discussion}

We introduced a new approach for model-based planning that learns to construct, evaluate, and execute plans.
Our continuous IBP represents the first fully learnable method for model-based planning in continuous control problems, using its model for both imagination-based planning and gradient-based policy optimization. Our results in the spaceship task show how the IBP's learned planning strategies can strike favorable trade-offs between external task loss and internal imagination costs, by sampling alternative imagined actions, chaining together sequences of imagined actions, and developing more complex imagination-based planning strategies.
Our results on a 2D maze-solving task illustrate how it can learn to build flexible imagination trees, which inform its plan construction process and improve overall performance.

In the future, the IBP should be applied to more diverse and natural decision-making problems, such as robotic control and higher-level problem solving. Other environment models should be explored, especially those which operate over raw sensory observations. The fact that the imagination is differentiable can be exploited for another form of model-based reasoning: online gradient-based control optimization.

Our work demonstrates that the core mechanisms for model-based planning can be learned from experience, including implicit and explicit domain knowledge, as well as flexible plan construction strategies. By implementing essential components of a planning system with powerful, neural network function approximators, the IBP helps realize the promise of model-based planning and opens new doors for learning solution strategies in challenging problem domains.

\bibliographystyle{plain}
\small

\bibliography{bibliography}

\newpage

\section*{Appendix: More About the Maze Tasks}

\begin{figure}[h]
\centering
\begin{tabular}{cccc}
    \includegraphics[width=0.20\textwidth]{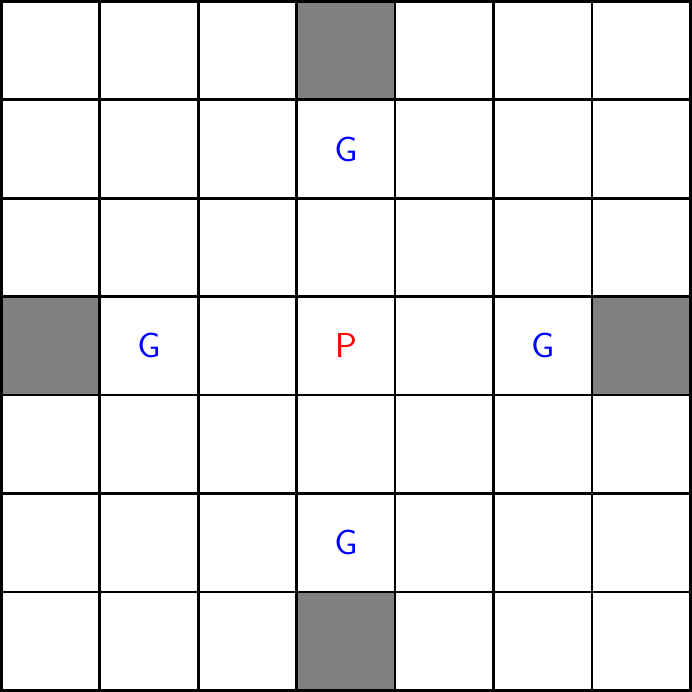} &
    \includegraphics[width=0.20\textwidth]{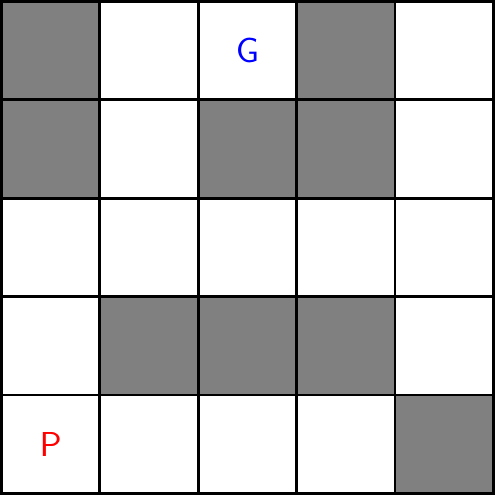} &
    \includegraphics[width=0.20\textwidth]{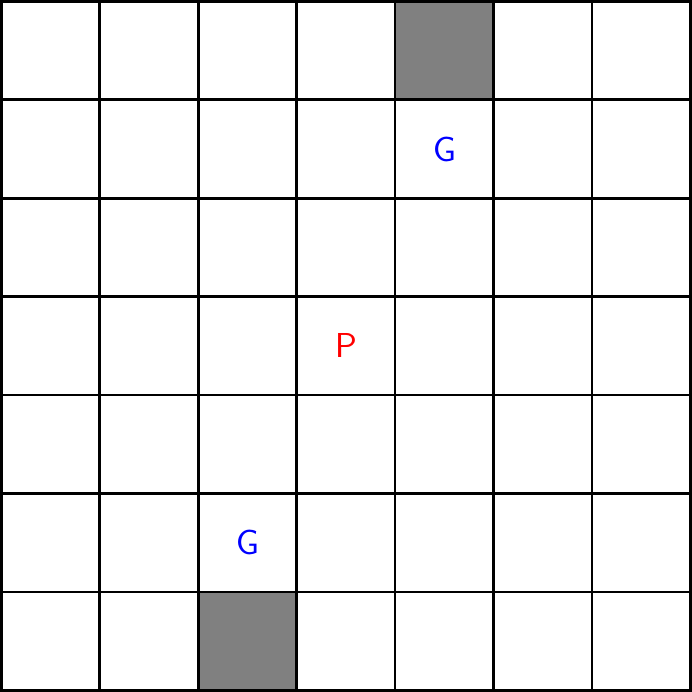} &
    \includegraphics[width=0.20\textwidth]{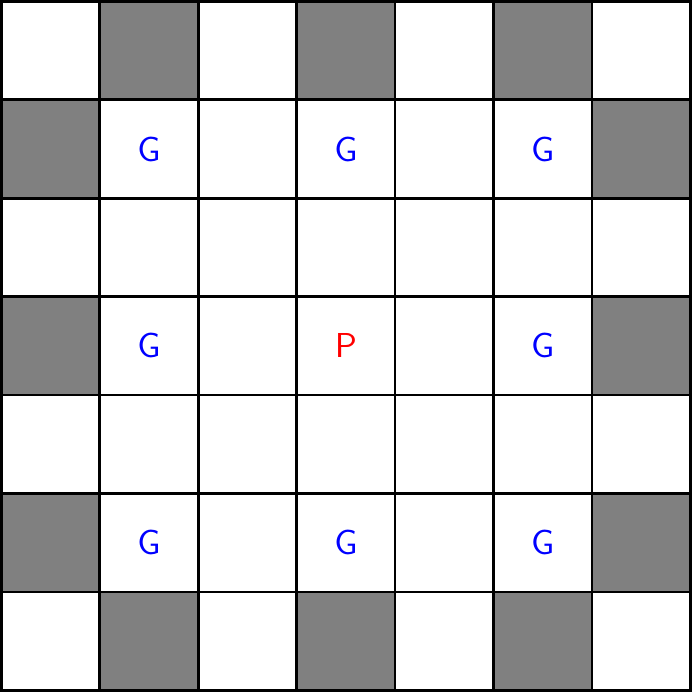}
\end{tabular}
\caption{Single maze, multiple goals setup.  All of them share the same layout, and the agent is forced to use the same set of Q-values for all goal positions.  The first three are training goal positions, and the last one is unseen during training and is used to test the ability of the agent to generalize.}
\label{fig:single-maze-multi-goals}
\end{figure}

\textbf{Single maze multiple goals:} The 4 mazes used in the single maze, multiple goals study are shown in Figure~\ref{fig:single-maze-multi-goals}.  According to the reward scheme introduced in the main paper, the optimal reward (normalized by 20, the episode length) for each of the 4 mazes are $\frac{-3+17}{20} = 0.8$, $\frac{-6+14}{20}=0.4$, $\frac{-7+13}{20}=0.3$ and $\frac{-7+13}{20}=0.3$.

For an agent that does not use imagination, the optimal strategy would be to go to the closest goal all the time.  However, given enough imagination steps, the agent would be able to disambiguate different environments, and act accordingly.

\begin{figure}[h]
  \centering
  \begin{tabular}{cccc}
      \includegraphics[width=0.20\textwidth]{imagine5_traces.pdf} &
      \includegraphics[width=0.20\textwidth]{imagine9_traces.pdf} &
      \includegraphics[width=0.20\textwidth]{imagine13_traces.pdf} &
      \includegraphics[width=0.20\textwidth]{imagine15_traces.pdf} \\
      \includegraphics[width=0.20\textwidth]{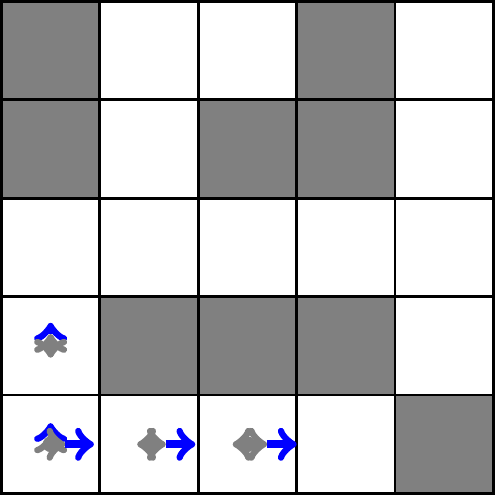} &
      \includegraphics[width=0.20\textwidth]{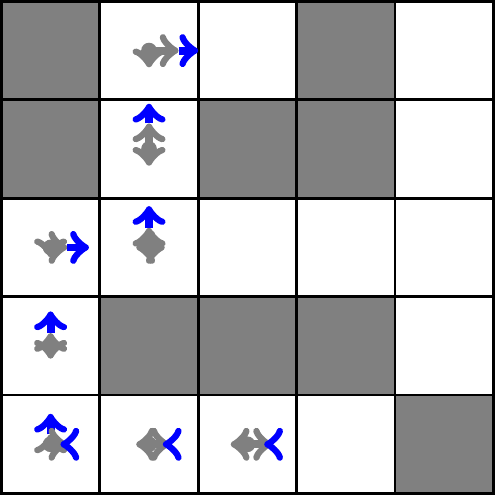} &
      \includegraphics[width=0.20\textwidth]{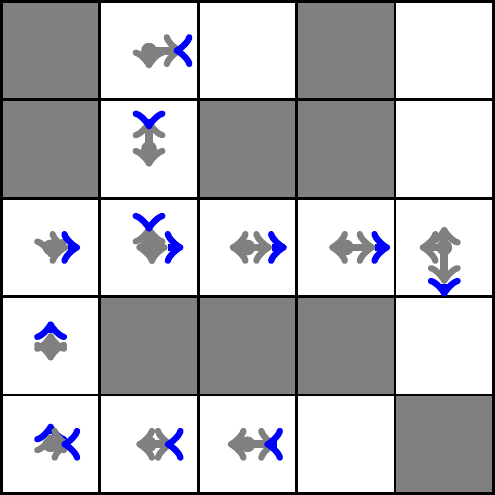} &
      \includegraphics[width=0.20\textwidth]{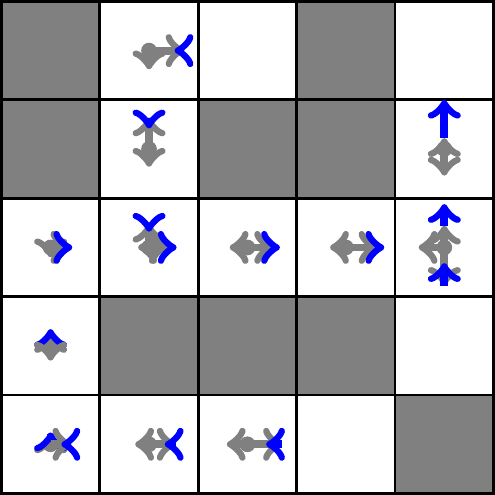}
  \end{tabular}
  \caption{Imagination rollouts with different imagination budget (first row), and the corresponding updates to the plan context $c$ (second row).  In the update plots, the gray arrows represent the original $Q+c$ values for each action before the update, and the blue arrows represent the update to $c$.}
  \label{fig:single-maze-multi-goals-imagine-and-update}
\end{figure}

Figure~\ref{fig:single-maze-multi-goals-imagine-and-update} shows the imagination rollouts along with the corresponding updates to the planning context $c$.  It is clear from the updates that imagination helps to adjust the agent's policy to adapt better to the particular episode.

\begin{figure}[h]
\centering
\begin{tabular}{cccc}
    \includegraphics[width=0.20\textwidth]{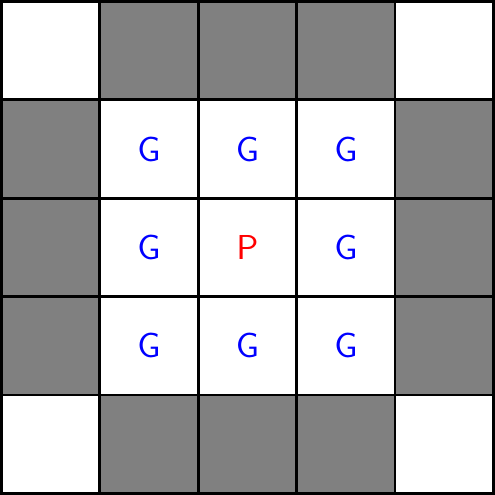} &
    \includegraphics[width=0.20\textwidth]{maze_group_1.pdf} &
    \includegraphics[width=0.20\textwidth]{maze_group_2.pdf} &
    \includegraphics[width=0.20\textwidth]{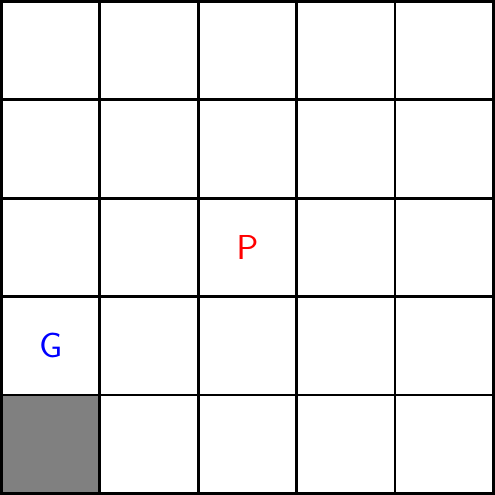} \\

    \includegraphics[width=0.20\textwidth]{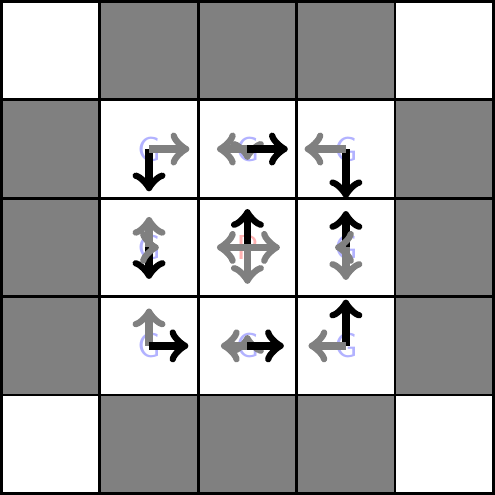} &
    \includegraphics[width=0.20\textwidth]{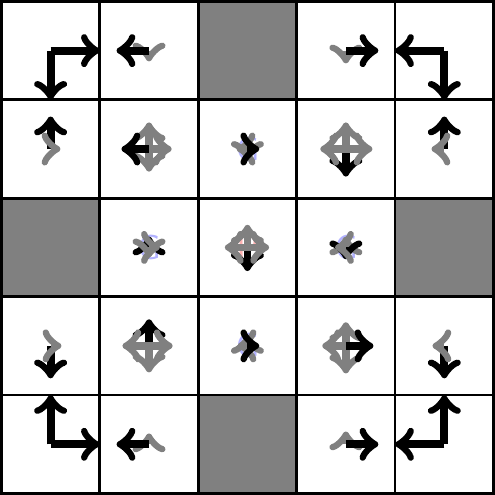} &
    \includegraphics[width=0.20\textwidth]{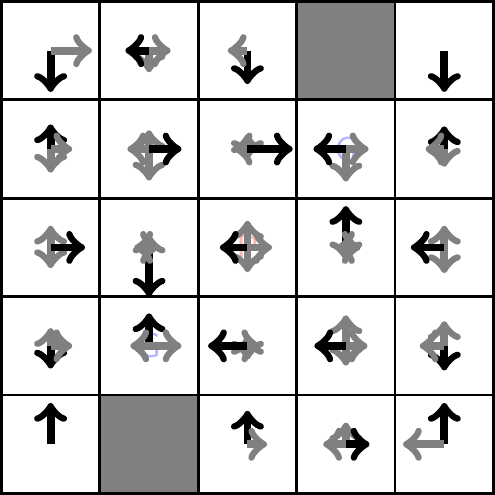} &
    \includegraphics[width=0.20\textwidth]{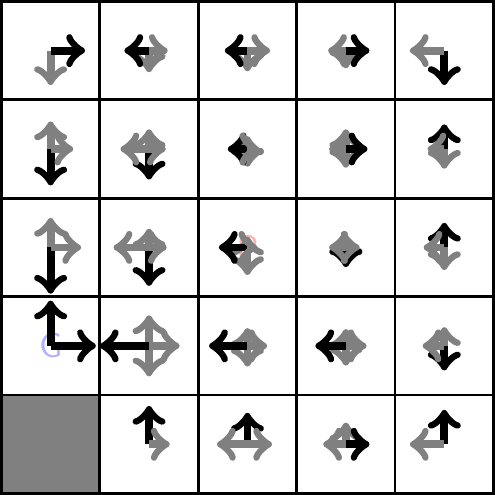}
\end{tabular}
\caption{Multiple mazes setup (top row) and the corresponding learned default policies (bottom row).  The Q-values are represented as gray arrows, and the Q-value for the action with the maximum value is highlighted as black solid arrows.}
\label{fig:multiple-mazes-and-default-policy}
\end{figure}

\textbf{Multiple mazes multiple goals:} Figure~\ref{fig:multiple-mazes-and-default-policy} shows the configuration of the 5$\times$5 mazes we used for this exploration.  Here each maze may have more than one goal positions, and the agent has one set of $Q$-values for each of the mazes.  Along with the maze configurations we are also showing the default policy learned for these mazes, which basically points to the closest possible goal location.

\begin{figure}
\centering
\begin{tabular}{cc}
    \includegraphics[width=0.45\textwidth]{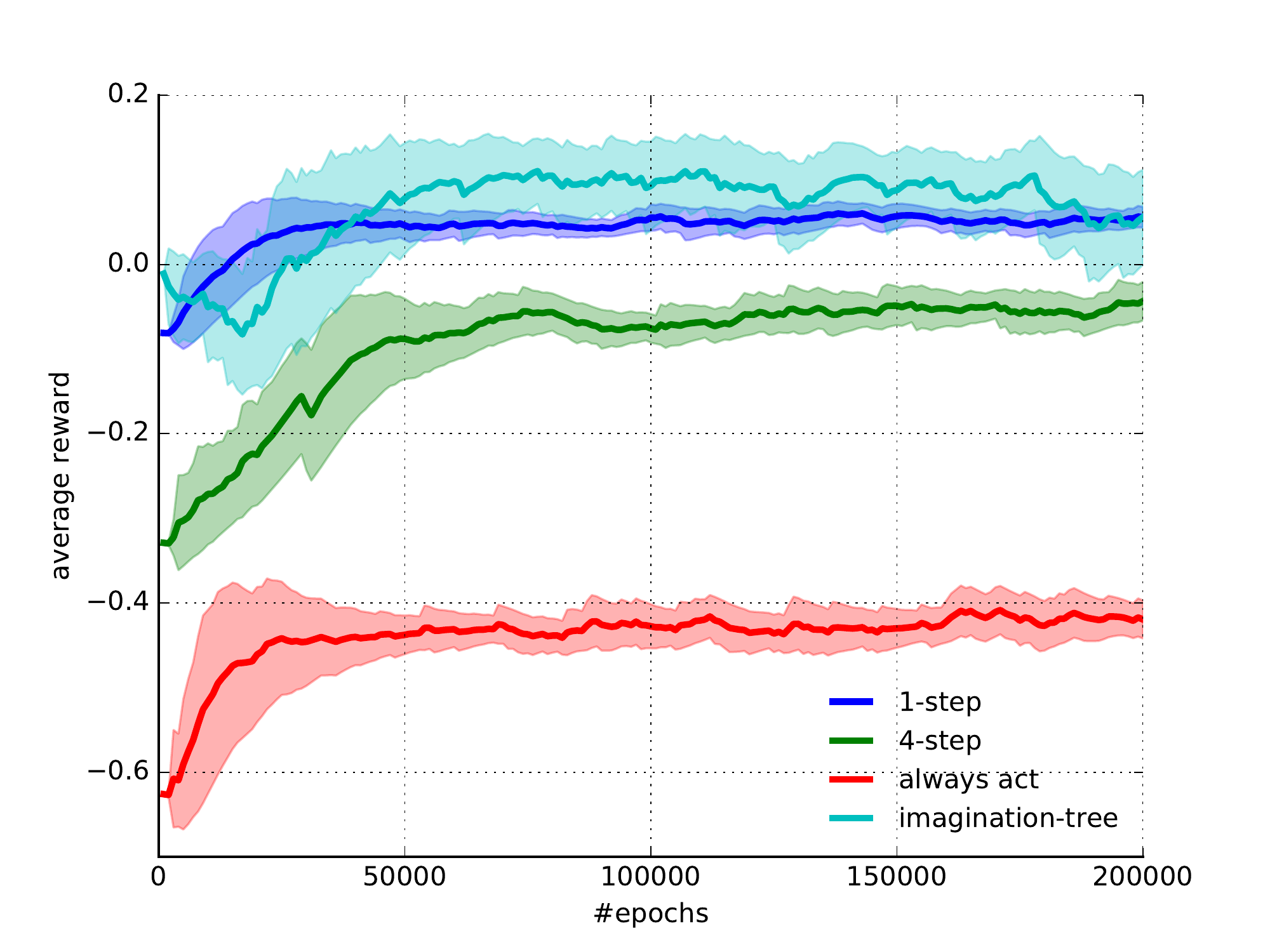} &
    \includegraphics[width=0.45\textwidth]{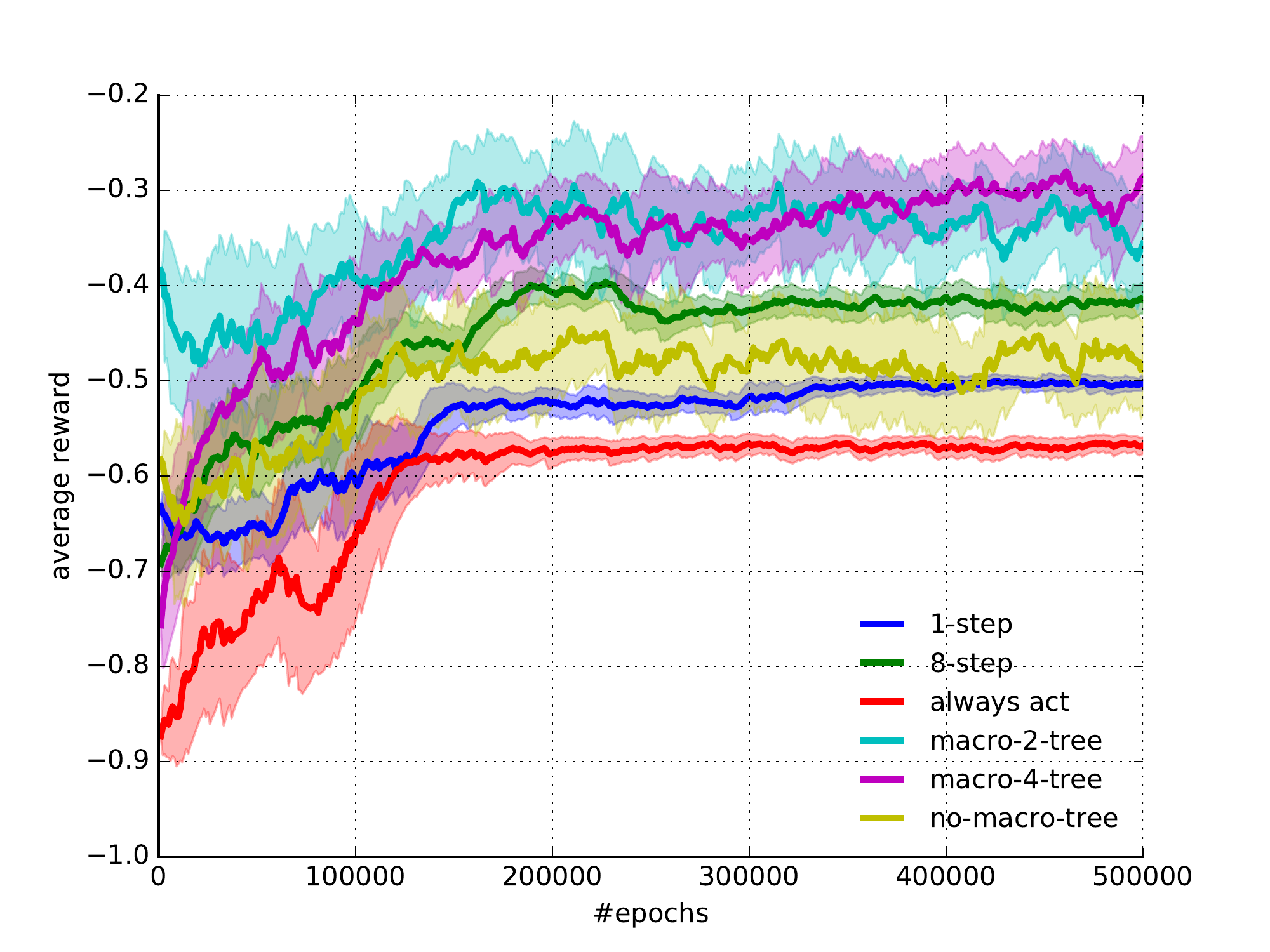}
\end{tabular}
\caption{Performance of different imagination strategies as the model gets learned.  Left: 5$\times$5 maze, all methods have the same imagination budget of 4 steps, right: 7$\times$7 maze, all methods have the same budget of 8 imagination steps.  To make the overall trend more visible, the solid lines show the exponential moving average and the shaded regions correspond to exponential moving standard deviation.  On the 7$\times$7 maze, it is beneficial to operate on 2 or 4-step ``macro actions'' instead of exploring simple actions only.}
\label{fig:maze-meta-controller-comparison}
\end{figure}

Figure~\ref{fig:maze-meta-controller-comparison} left shows the comparison between different imagination strategies, where we compare the learned neural net imagination strategy with fixed strategies including 1-step, $n$-step ($n=4$) and no imagination.  For this comparison, all the imagination based methods have the same imagination budget at each step.  But the learned strategy can adapt better to different scenarios.

\begin{figure}[h]
\centering
\begin{tabular}{cccc}
    \includegraphics[width=0.22\textwidth]{maze0.pdf} &
    \includegraphics[width=0.22\textwidth]{maze1.pdf} &
    \includegraphics[width=0.22\textwidth]{maze3.pdf} &
    \includegraphics[width=0.22\textwidth]{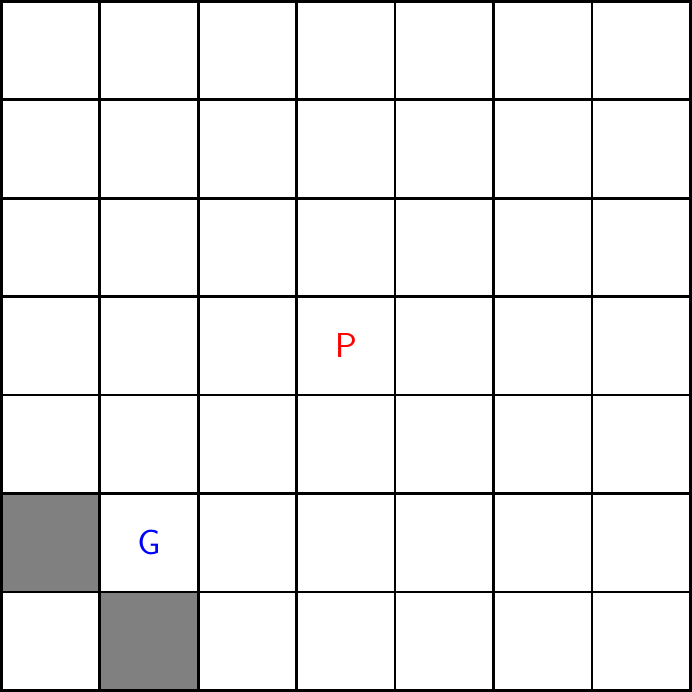}
\end{tabular}
\caption{7$\times$7 mazes, a scaled up version of the 5$\times$5 mazes used previously.  This is more challenging than the 5$\times$5 mazes and poses more challenges to the imagination process.}
\label{fig:7x7-mazes}
\end{figure}

Figure~\ref{fig:7x7-mazes} show the 7$\times$7 mazes we used as a scaled up version of the 5$\times$5 mazes.  These are more challenging as the agent needs to explore in a much bigger space to get useful information.  We proposed to use ``macro-actions'' using rollouts of length more than 1 as basic units.  Figure~\ref{fig:maze-meta-controller-comparison} right shows the comparison between different imagination strategies for this task.  Using macro-actions consistently improves performance, while not using ``macro-actions'' does not perform as well, potentially because this is a much harder learning problem, in particular, the action sequences are much longer, and the useful reward information are delayed even further.

\section*{Appendix: More About the Spaceship Task}

The setup we used followed closely that of \cite{hamrick2017metacontrol}, and most details are identical
with the ones described in that paper. To re-iterate, we
 used Adam\cite{kingma2014adam} as the optimizer, where the learning rate was decreased by 5\% percent
 every time the validation error increased. Validation error was
 computed every $1000$ iterations. Each iteration consists of evaluating the gradient over a minibatch of $1000$ episodes. We used
 different initial learning rates for the different components. For training the model of the world we used a learning rate of $0.001$,
 for the controller $0.0003$ and for the manager $0.0001$. We rely on gradient clipping, with the max gradient norm allowed of $10$. The agents are trained for a total of $100,000$ iterations. The pseudocode of our most complex agent is given below:

\begin{algorithm}[t]
\begin{algorithmic}[1]
\Function{$a^M$}{$x, x^*$}
  \State $h\gets ()$\Comment{Initial empty history}
  \State $n_{real}\gets 0$
  \State $n_{imagined} \gets 0$
  \State $x_{real} \gets x$
  \State $x_{imagined} \gets x$
  \While {$n_{real} < n_{max-real-steps}$}\Comment{$n_{max-real-steps}$ is a hyper-parameter}
      \State $r \gets \pi^M(x_{real}, x^*, h_n)$
      \If {$r==0$ or $n_{imagined} < n_{max-imagined-steps}$} \Comment{Execute action}
      \State $c \gets \pi^C(x_{real}, x^*, h_n)$
      \State $x_{real} \gets World(x_{real}, c)$ \Comment{Execute control}
      \State $n_{real} += 1 $ \Comment{Increment number of executed actions}
      \State $n_{imagined} = 0$
      \State $x_{imagined} \gets x_{real} $ \Comment{Reset imagined state}
      \ElsIf {$r==1$}
      \State $c \gets \pi^C(x_{real}, x^*, h_n)$
      \State $x_{imagined} \gets I(x_{real}, c)$ \Comment{Imagine control from real state}
      \State $n_{imagined} += 1 $ \Comment{Increment number of executed actions}
      \ElsIf {$r==2$}
      \State $c \gets \pi^C(x_{imagined}, x^*, h_n)$
      \State $x_{imagined} \gets I(x_{imagined}, c)$ \Comment{Imagine control from previous imagined state}
      \State $n_{imagined} += 1 $ \Comment{Increment number of executed actions}
      \EndIf
      \State $h\gets \mu(h, c, r, x_{real},x_{imagined}, n_{real}, n_{imagined})$\Comment{Update the history}

  \EndWhile
\EndFunction
\end{algorithmic}
\caption{IBP agent with our simple imagination strategy. $x$ is the current scene and $x^*$ is the target. $h$ is the current context}
\label{alg:metacontroller}
\end{algorithm}

\end{document}